\documentclass[preprint,12pt,authoryear]{elsarticle}

\usepackage[para,online,flushleft]{threeparttable}
\usepackage{booktabs} 
\usepackage{multirow} 

\usepackage{amssymb}
\usepackage{amsmath}
\usepackage{hyperref}

\journal{Nuclear Physics B}

\begin{document}

\begin{frontmatter}



\title{MRI-Based Brain Age Estimation with Supervised Contrastive Learning of Continuous Representation}


\author{Simon Joseph Clément Crête, Marta Kersten-Oertel, Yiming Xiao} 

\affiliation{organization={Department of Computer Science and Software Engineering, \\Concordia University},
            city={Montréal},
            country={Canada}}

\begin{abstract}
MRI-based brain age estimation models aim to assess a subject's biological brain age based on information, such as neuroanatomical features. Various factors, including neurodegenerative diseases, can accelerate brain aging and measuring this phenomena could serve as a potential biomarker for clinical applications. While deep learning (DL)-based regression has recently attracted major attention, existing approaches often fail to capture the continuous nature of neuromorphological changes, potentially resulting in sub-optimal feature representation and results. To address this, we propose to use supervised contrastive learning with the recent Rank-N-Contrast (RNC) loss to estimate brain age  based on widely used T1w structural MRI for the first time and leverage Grad-RAM to visually explain regression results. Experiments show that our proposed method achieves a mean absolute error (MAE) of 4.27 years and an $R^2$ of 0.93 with a limited dataset of training samples, significantly outperforming conventional deep regression with the same ResNet backbone while performing better or comparably with the state-of-the-art methods with significantly larger training data. Furthermore, Grad-RAM revealed more nuanced features related to age regression with the RNC loss than conventional deep regression. As an exploratory study, we employed the proposed method to estimate the gap between the biological and chronological brain ages in Alzheimer's Disease and Parkinson's disease patients, and revealed the correlation between the brain age gap and disease severity, demonstrating its potential as a biomarker in neurodegenerative disorders.
\end{abstract}



\begin{keyword}
Brain aging, structural MRI, contrastive learning, deep learning, neuro-degeneration


\end{keyword}

\end{frontmatter}



\section{Introduction}
Due to disease, genetics, and environmental factors \citep{Franke2019TenYears,franke2013}, biological aging of the human body can deviate from the normal aging process that is commonly documented by the subject's chronological age. Precise measurement of biological aging offers a more meaningful metric for clinical applications \citep{aging:2020}, and is particularly useful for studying the human brain, which exhibits significant anatomical, biochemical, and functional changes with age \citep{Peters:2006, Fujita:2023}. Furthermore, most neurodegenerative diseases are characterized by neuroanatomical changes similar to advanced aging \citep{lifespanChangesCoupe, GMchangePD, COLEreview:2017}. Thus, the difference between  estimated biological and chronological ages of the brain (i.e., brain age gap) may provide a valuable biomarker and enable better management and understanding of neurological diseases \citep{cole:2016}. However, methods to accurately estimate the biological age of the brain still require further investigation.

Over the last decade, significant work has been produced to create computational models for brain age estimation using neuroimaging data \citep{Franke2019TenYears}. With wide adoption in the clinic and great anatomical details, T1w structural MRIs have become the main data type used for the task \citep{mishra2023}, with machine learning (ML) and deep learning (DL) approaches able to produce brain age estimation at varying accuracy \citep{brainagepredRev}. Typically, a brain age estimation model will be trained on data from healthy subjects, assuming that their biological and chronological ages are consistent. In earlier approaches, traditional ML regression algorithms employed engineered anatomical features extracted from MRIs, such as local cortical thickness and volumetric measurements. For instance, \cite{Franke2010} used relevance vector regression on segmented gray matter volume and achieved a mean absolute error (MAE) of 5 years. \cite{liem2016} used linear support vector regression models on multiple features, including cortical thickness, cortical surface area, and subcortical volume measurements to achieve an MAE of 4.29 years. These feature-based methods rely heavily on image post-processing algorithms, whose accuracy and robustness can vary and often require relatively long computational time. 

More recent DL techniques for MRI-based brain age estimation allow minimal data processing and fast inference time (e.g., in a fraction of a second), thus making them more attractive than the traditional ML-based approaches in deployment. So far, many DL algorithms for the designated task have been proposed and led to significant performance improvements \citep{brainagepredRev}. As one of the earliest DL approaches, \cite{cole:2016} demonstrated that 3D-convolutional neural networks (CNNs) trained on structural MRIs could predict brain age with an MAE of $\sim$ 4 years with an $R^2$ of 0.94, outperforming most traditional ML methods at the time. Later, advancements in CNN architectures have further improved the state-of-the-art (SOTA) accuracy. \cite{Yin:2023} achieved an MAE of 2.3 years with a customized 3D CNN while \cite{pyment} and \cite{SFCN} achieved 3.90 years and 2.14 years, respectively with the VGG-like SFCN architectures. With Residual network (ResNet) architectures, MAEs of 2.70$\sim$3.39 years \citep{jonsson2019,Ning2021} have also been achieved. \cite{GlobLocTransformer} presented a global-local approach with ResNet backbones, coupled with a Transformer to reach an MAE of 2.7 years for a population with an age range of 0$\sim$97 years old (yo). Leveraging 3D U-Nets to estimate voxel-wise age maps, \cite{Nguyen:2024} achieved an MAE of 3.83 years on an aging cohort and 1.91 years on a young one. Lastly, with the rise of contrastive representation learning, \cite{multisitecon} adopted this training technique with a custom Y-aware regression loss to reach an MAE of 2.55 years on a dataset of mostly young subjects using gray matter volumes from voxel-based morphometry (VBM) analysis. However, despite their reported accuracy, as these existing methods employ different training and testing datasets, to fairly assess and compare their performance, a common benchmark should be necessary, and notably they typically require a large amount of training samples to achieve satisfactory results. More importantly, these prior deep regression methods rely on imposing constraints on the final prediction in an end-to-end manner without explicitly emphasizing a regression-aware presentation. As demonstrated by \cite{zha2023rankncontrast} with natural images, this can often lead to fragmented deep feature representation in the manifold space and thus fail to capture the intrinsic continuous relationship, such as the structural changes shown in healthy brain MRIs with the growing age, resulting in sub-optimal performance, training efficiency, and potentially application-relevant algorithm explainability. 

With structural MRI-based brain age estimation techniques, studies \citep{mishra2023} have suggested that the severity of certain neurological conditions can potentially be quantified, with the observation that the mean brain age gap between the estimated biological and chronological ages of a patient population is typically higher than a healthy one. To date, a few groups have reported this in leading neurodegenerative conditions, such as Alzheimer's disesase (AD) and Parkinson's disease (PD) \citep{Sendi:2021, Eickhoff:2021, Nguyen:2024}. In addition, studies using ML on processed MRIs have found weak to moderate correlations between brain age gap and AD severity metrics \citep{lowe2016, beheshti2018, Yin:2023}. For PD, limited brain age estimation studies have found similar types of correlations, with only weak ones being reported \citep{Eickhoff:2021, chen2024}. While these findings are promising, there is still no consensus on the utility of brain age as a biomarker for neurodegenerative diseases \citep{Franke2019TenYears,accvsutility}. Further investigations are still required, especially with different DL approaches.

To address the drawbacks of sub-optimal feature representation learning and heavy reliance on large training dataset in conventional DL regression methods for structural MRI-based brain age estimation, we propose to leverage the latest contrastive learning (CL) paradigm, which has demonstrated superior performance compared to typical end-to-end training in a wide range of applications \citep{supconlossgoogle,zha2023rankncontrast, contrastiveReview,contrastiveSegmen}. Specifically, for the popular ResNet model, we investigate the adoption of supervised contrastive learning with Rank-N-Contrast (RNC) loss \citep{zha2023rankncontrast}, which was used for facial age prediction and in our task, can explicitly enforce continuous anatomical feature representation from T1w MRIs according to their ranking of age. We compare our proposed method against conventional end-to-end regression ResNets, as well as two popular publicly available state-of-the-art methods, Simple Fully Convolutional Neural Network on regression (SFCN-reg) \citep{pyment} and Brain Structure Ages (BSA) \citep{Nguyen:2024}, which were trained with substantially larger datasets and represent two primary types of computational approaches for the designated task. To help gain insights into the decision-making of the proposed technique and demonstrate the advantage of the RNC loss function, we employed the gradient-weighted regression activation mapping (Grad-RAM) technique, which is an adaptation of gradient-weighted class activation mapping (Grad-CAM) \citep{gradcam} for regression tasks, to obtain 3D activation maps that visualize the key regions related to the model outputs. Finally, to explore the applicability of our proposed method in downstream clinical applications, we explore the correlations between the estimated brain age gap from our model and clinical assessments for the severity of AD and PD population. To the best of our knowledge, our study is the first to employ contrastive learning with RNC loss for 3D medical data, particularly for the task of structural MRI-based age estimation. We demonstrate its advantages in enhancing performance with limited training data while better capturing relevant anatomical regions to reflect brain aging than conventional end-to-end training. 


\section{Methods and Materials}
We propose to employ the RNC contrastive learning technique for the first time to enhance the performance of MRI-based age regression. Considering the prior success and popularity in brain age estimation \citep{DLbrainagepredRev, jonsson2019, ZHANG2024}, we adopt the ResNet models \citep{resnet} as the backbone of our proposed method with the contrastive RNC loss. To ensure the optimal architecture and training configurations of the network's backbone, we first inspect the impacts of ResNet depth, data augmentation, and image resolution on age regression accuracy by using conventional end-to-end training. Upon selecting the optimal ResNet configuration from all the variants with the consideration of accuracy and computational constraints, we apply supervised contrastive learning and the results are compared against two state-of-the-art techniques, as well as different ResNet variants with end-to-end training. In addition, we implement the Grad-RAM technique to provide visual saliency maps for the DL-based age estimation results to reveal insights regarding brain aging from the proposed method in comparison to the end-to-end training. Finally, we examine the application of our proposed method to compute the brain age gap based on an AD and PD cohort, which were correlated with disease severity metrics as a potential biomarker for the diagnosis of neurodegnerative disorders. All models were trained with 4 Nvidia A100-40Gb GPUs courtesy of Digital Research Alliance of Canada.

\subsection{Dataset}
\begin{figure}[bt]
    \centering
    \includegraphics[width=0.85\linewidth]{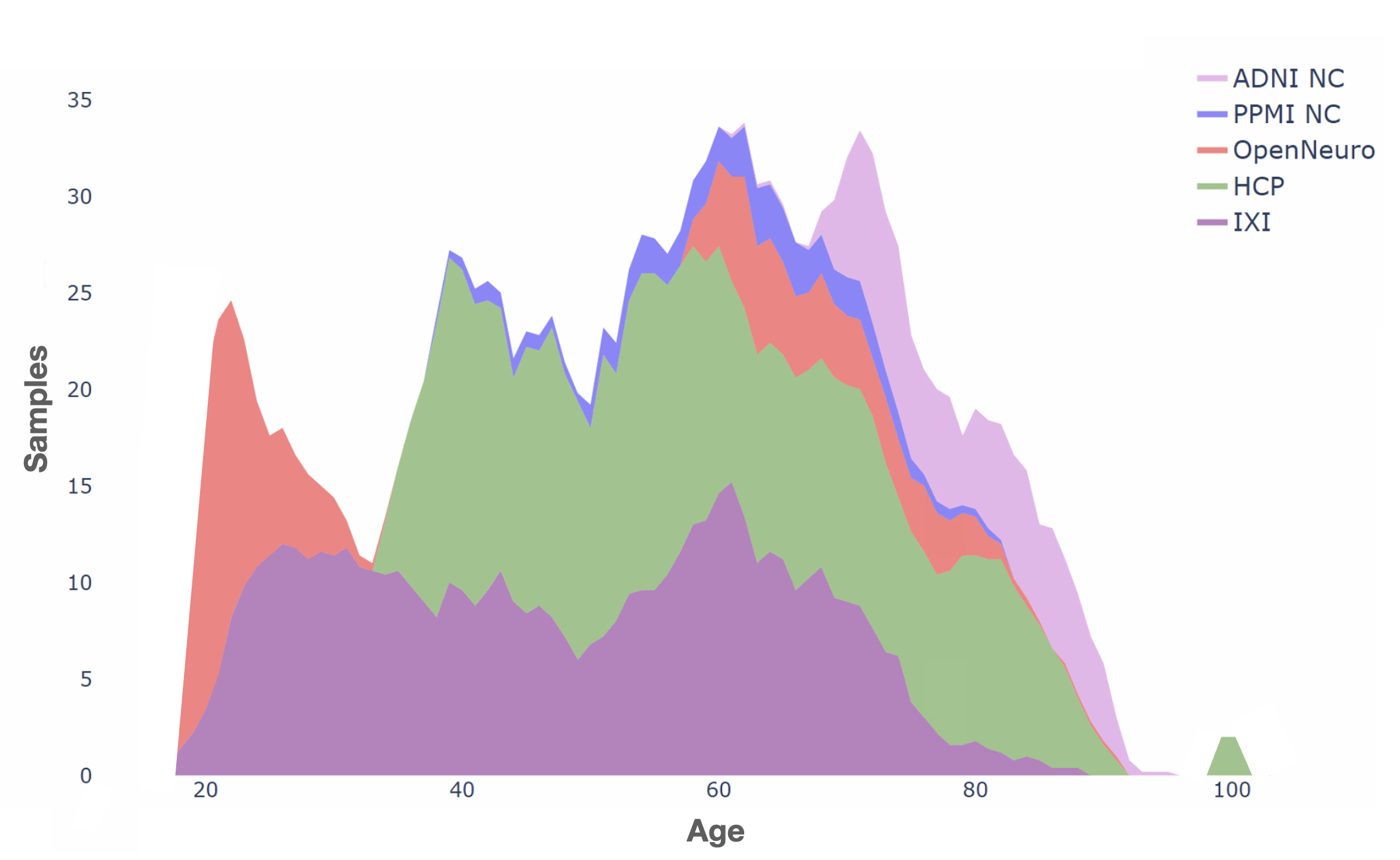}
    \caption{Relative age distribution of cognitively normal subjects across datasets (n = 1618).}
    \label{fig:ageDistr}
\end{figure}

We collected a total of 1,618 healthy subjects' 3T T1w brain MRIs with the corresponding chronological age and sex (age=52.8±18.5 yo, range=20$\sim$100 yo, 872 females) for the development and validation of the proposed method from five publicly available datasets, including the Human Connectome Project Aging (HCP) \citep{HCP}, Alzheimer’s Disease Neuroimaging Initiative (ADNI) \citep{ADNI}, IXI Dataset (\url {https://brain-development.org/ixi-dataset/}), OpenNeuro (Neurocognitive aging data release with behavioral structural and multi-echo functional MRI measures \citep{openneuro}), and Parkinson's Progressive Markers Initiative (PPMI) \citep{ppmi}. The details for each dataset are summarized in Table \ref{table:data} while the distributions of the datasets are depicted in Fig. \ref{fig:ageDistr}.

We used a train/validation/test data split with 80\%/10\%/10\% of the full data respectively for all model training and validation. More specifically, the test dataset (149 subjects with 79 females, age=51.26±19.32 yo) have 41, 46, 47, and 15 subjects in the 20-40 yo, 40-60 yo, and 60-80 yo, and 80+ yo age groups, respectively. To further explore the potential application of the proposed method in gauging progression of neurodegenerative conditions, we also collected 61 Alzheimer's disease patients (age=78.1±7.3 yo, 22 female) with ADAS-cog 11 score of 25.56±9.46 from the ADNI database, and 80 Parkinson's disease patients (age=66.8±9.1yo, 24 female) with the UPDRSIII and H\&Y scores of 27.91±13.88 and 2.28±0.58 from the PPMI dataset, respectively. Specifically, for the AD patients, we selected those that were scanned 24 months after their baseline visits, and PD patients's data 48 months after baseline visits were selected. For both patient cohorts, we ensure a variety of mild and severe disease cases, and a variety of age ranges.

\begin{table}[bt]
\caption{Statistics of datasets used for normal control (NC), Alzheimer's Disease (AD), and Parkinson's Disease (PD). NC = Normal Control; AD = Alzheimer's Disease; PD = Parkinson's Disease; IXI = Information eXtraction from Images; HCP = Human Connectome Project; ADNI = Alzheimer's Disease Neuroimaging Initiative; PPMI = Parkinson's Progression Markers Initiative.}
\label{table:data}
\centering
\begin{threeparttable}
\setlength{\tabcolsep}{6pt} 
\renewcommand{\arraystretch}{1} 

\resizebox{\textwidth}{!}{
\begin{tabular}{lcccc}
\toprule
\textbf{Dataset} & \textbf{\boldmath{$N_{\text{samples}}$}} & \textbf{Age:Mean±Std (yo)} & \textbf{Age Range (yo)} & \textbf{Sex(F/M)} \\ 
\midrule
\multicolumn{5}{l}{\textbf{Normal Control (NC) Datasets}} \\
IXI & 562 & $48.7 \pm 16.5$ & 20--86.3 & 311 / 251 \\
HCP & 635 & $59.8 \pm 15.5$ & 20--100 & 355 / 280 \\
OpenNeuro & 223 & $41.0 \pm 23.3$ & 20--89 & 129 / 94 \\
ADNI NC & 136 & $78.0 \pm 6.5$ & 69--93 & 55 / 81 \\
PPMI NC & 62 & $61.4 \pm 9.8$ & 40.2--81.1 & 22 / 40 \\
\textbf{NC Totals} & 1618 & $\boldsymbol{52.8 \pm 18.5}$ & \textbf{20--100} & \textbf{872 / 746} \\
\midrule
\multicolumn{5}{l}{\textbf{Patient Datasets}} \\
ADNI AD & 61 & $78.1 \pm 7.3$ & 59--91 & 22 / 39 \\
PPMI PD & 80 & $66.8 \pm 9.1$ & 43--86 & 24 / 56 \\
\bottomrule
\end{tabular}}

\end{threeparttable}
\end{table}

\subsection{Image Preprocessing}
All collected T1w MRIs were preprocessed with the MINC Toolkit (https://bic-mni.github.io/) to ensure data quality and facilitate image analysis. Specifically, we obtained the brain mask with the BEaST algorithm \citep{beast} and performed N4 non-uniformity correction \citep{n4bias} for each scan. Afterwards, the skull-stripped MRIs were registered and resampled to the MNI-ICBM152 space with a 12-parameter affine transformation using normalized cross-correlation (NCC) as the cost function and B-spline interpolation. The final images are of dimension 193x229x193 voxels with 1x1x1 $mm^{3}$ resolution. To mitigate computational complexity, we also generated a downsampled version of the same dataset at 2x2x2 $mm^{3}$ resolution and size of 96x114x96 voxels. A visual quality control was performed on each sample by the first author SC. 

\subsection{Data Augmentation}
To improve training robustness and mitigate overfitting, we adopted four data augmentation techniques for the training dataset, each with a 50\% probability of being applied, including 1) random translation of ±10 voxels in each cardinal direction, 2) random rotation of 0.1 to 0.5 radians in any direction, 3) random Gaussian noise addition with 0 mean and 0.025 standard deviation, and 4) random image cropping with the size no smaller than 70\% of the original image (resized to the original image dimensions).

\subsection{Deep Learning Models and Experiments}

\subsubsection{3D ResNet Backbone Evaluation}
We evaluated 3D ResNets \citep{resnet} with varying depths, including 18, 50, and 101, to help determine the optimal backbone configuration for our intended contrastive learning approach, while the rest also serve as the baseline models in a end-to-end training manner for performance assessment. Each variant of the ResNet architecture underwent training for 300 epochs with early stopping to regress age estimates based on preprocessed T1w brain MRI scans. MAE (or L1 loss) was used as the loss function with the Adam optimizer and a learning rate of 0.001. For different ResNet depths, we further implemented versions with high-resolution and low-resolution image inputs, resulting in six different variants. For variants that use low resolution images of 2x2x2 mm³, a batch size of 8 was used, while for 1x1x1 mm³ resolution counterparts, a batch size of 4 was employed. We conducted experiments with both augmented and un-augmented training datasets to assess the impact on accuracy improvement. The accuracy of different ResNet variants was compared.

\subsubsection{Contrastive Learning with Rank-N-Contrast}
Contrastive learning (CL) has recently emerged as a powerful DL technique for efficient and robust feature representation learning. The main idea is to compare an anchor sample against others in a batch, so that the training process can maximize the distance measures between similar data points (positive pairs) and minimize those between dissimilar ones (negative pairs) in the feature space. However, existing CL methods are primarily designed for image classification, and very few investigated tailored formulation for regression tasks. To address this, \cite{zha2023rankncontrast} recently proposed the Rank-N-Contrast (RNC) method, which has outperformed SOTA methods in various image-based attribute regression tasks. Different from conventional contrastive learning methods, where sample similarity is defined based on their categories, the Rank-N-Contrast technique is specifically designed to model continuous feature representation by structuring the learning process to respect the ranking of the images' numerical labels. So far, it has shown excellent results in natural vision domains, including age regression from facial photographs, a problem that is akin to our intended application. Since aging-related anatomical changes occur continuously across the life-span, the benefits of RNC demonstrated in natural images in terms of accuracy gain and training efficiency motivate its adaptation to MRI-based brain age estimation. For our designated task, the RNC method will define brain MRI scans with closer age labels as positive pairs while those with more distant age labels as negative pairs. We briefly elaborate the formulation of RNC method in the context of our application below, and full details can be found in the original paper \citep{zha2023rankncontrast}.

For a given anchor MRI $I_i$ and the associated age $y_i$, its deep feature embedding through a DL network is represented by $v_i$. In a training batch of $2N$ subjects, for any other sample $\{v_j,y_j\}$, we denote the set of samples that have bigger age difference to the anchor than $v_j$ as $S_{i,j} = \{v_k | k \neq i, |y_i-y_k|\geq|y_i-y_j|\}$. To effectively enforce ordered relationships in the learned feature embeddings, the RNC loss $L_{\text{RNC}}$ is defined as the average negative log likelihood of $v_j$ over all other samples in a given batch. Therefor, the per-sample RNC loss is given by:

\[l^{(i)}_{\text{RNC}} = \frac{1}{2N - 1} \sum_{\substack{j=1 \\ j \neq i}}^{2N} -log (P(v_j|v_i,S_{i,j}))= \frac{1}{2N - 1} \sum_{\substack{j=1 \\ j \neq i}}^{2N} - \log \frac{\exp(\text{sim}(\mathbf{v}_i, \mathbf{v}_j)/\tau)}{\sum\limits_{\mathbf{v}_k \in S_{i,j}} \exp(\text{sim}(\mathbf{v}_i, \mathbf{v}_k)/\tau)}
\]
where \( \text{sim}(\mathbf{v}_i, \mathbf{v}_j) \) represents the similarity function between embeddings, and \( \tau \) is a temperature scaling factor. To enforce that the entire feature embeddings is ordered according to ranking orders in the label space, the loss is enumerated over all 2N samples $\{v_i\}$ as anchors:

\[
\mathcal{L}_{\text{RNC}} = \frac{1}{2N} \sum_{i=1}^{2N} l^{(i)}_{\text{RNC}} =\frac{1}{2N} \sum_{i=1}^{2N} \frac{1}{2N - 1} \sum_{\substack{j=1 \\ j \neq i}}^{2N} - \log \frac{\exp(\text{sim}(\mathbf{v}_i, \mathbf{v}_j)/\tau)}{\sum\limits_{\mathbf{v}_k \in S_{i,j}} \exp(\text{sim}(\mathbf{v}_i, \mathbf{v}_k)/\tau)}
\]

For encoding the input MRI as deep features, we adopted the 3D ResNet. Among all the ResNet backbone candidates mentioned in the previous section, ResNet50 with the data augmentation offers the best accuracy on average for end-to-end training under both high- and low-resolution settings, and it is followed by ResNet18 with data augmentation. Because in contrastive learning, sufficient pairs of comparisons of the anchor image against the positive and negative samples in a batch are necessary to allow optimal outcomes, batch size is a crucial parameter and should be maximized within the constraint of the GPU memory capacity. This is especially true for memory-intensive 3D MRI data and larger DL models. Therefore, for 2x2x2 $mm^{3}$ resolution data, a ResNet-50 backbone with batch size of 48 was used and we name the resulting model \textbf{RNC-lowRez} while for 1x1x1 $mm^{3}$ resolution data, a ResNet-18 backbone with a batch size of 24 was employed and we name the model \textbf{RNC-highRez}. 

\begin{figure}[bt]
    \centering
    \includegraphics[width=1\linewidth]{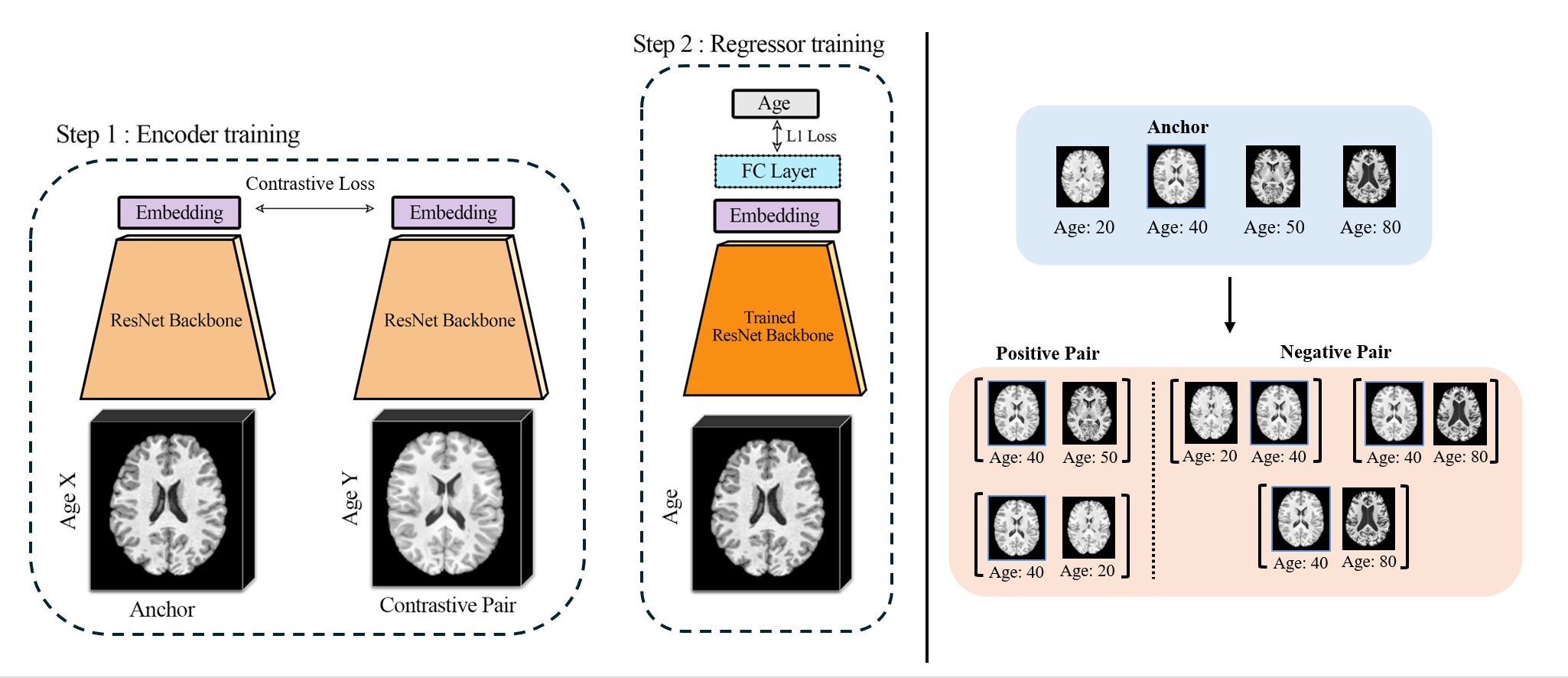}
    \caption{Left: Overview of the training process for the RNC  supervised learning model, with the first step consisting of training the ResNet backbone with the RNC loss (contrastive), and the second step training a FC regression layer with MAE/L1 loss. Right: Overview of the RNC loss \citep{zha2023rankncontrast} with a sample batch of data (n=4 for demonstration) and their labels (in blue box). Two example positive pairs and their corresponding negative pair(s) are shown (in orange box) when the anchor is the 40 year old brain MRI sample. When the anchor forms a positive pair with a 50 year old sample, their label distance is 10, hence the corresponding negative samples are the 20 year old and the 80 year old samples, whose label distances to the anchor are greater than 10. When the 20 year old sample creates a positive pair with the anchor, the 80 year old sample is a negative sample as it has a greater label distance to the anchor. }
    \label{fig:SupConTrain}
\end{figure}

To further inspect the impact of the batch size, we experimented with the batch size of 8, 16, and 24 for RNC-highRez, and the batch size of 16, 24, 32, and 48 for RNC-lowRez. Note that ResNet-101 was not considered due to computational and memory complexity requirements. For contrastive learning, our model training hyperparameters included a learning rate of 0.5, decay rate of 0.1, and momentum of 0.9 with the stochastic gradient descent (SGD) optimizer. The backbone ResNet encoders were trained for 1,000 epochs with the RNC loss. Afterwards, a regression head is added to the last layer of the backbone model, consisting of a fully connected (FC) layer to estimate the age from the encoded image features. In the last step, the backbone encoder weights were frozen and an MAE (i.e., L1) loss was used to train the FC layer, with the same training data for 100 epochs using the SGD optimizer (learning rate = 0.05, decay rate = 0.2, and momentum = 0.9). Figure \ref{fig:SupConTrain} shows an overview of the training steps.

\subsubsection{Comparison to the State-of-the-Art Methods}
We compared our contrastive learning technique for MRI-based brain age estimation against two popular pretrained public DL models, including the Simple Fully Convolutional Network with regression output (SFCN-reg) \citep{pyment} and BrainStructureAges (BSA) \citep{Nguyen:2024}, on the same test dataset. To best reflect the accuracy of these models, we decided not to train these models from scratch based on our curated dataset, but directly used their pretrained weights. Furthermore, to mitigate the impact of domain shift that can degrade performance due to differences in image pre-processing, we reprocessed our test data as instructed by each algorithm.  

 The SFCN-reg model \citep{pyment} was trained on a large and diverse neuroimaging dataset consisting of 42,829 T1w MRI scans from 21 public datasets, covering ages from 3 to 95 years, where HCP is the only dataset used in both our method and SFCN-reg. This is about 42 times of our training data size. In short, the SFCN-reg's architecture consists of a VGG-like network with five repeated convolutional blocks, each including a three-dimensional convolutional layer, batch normalization, ReLU activation, and max-pooling followed by a global average pooling layer, and finally a single output predicting age. It achieves an MAE of 2.47 years on an in-domain test set and 3.90 years on an external dataset, making it one of the best-performing models in the literature. For a fair comparison, our raw test dataset was preprocessed using the preprocessing script published by the authors of SFCN-reg on GitHub \citep{pyment}, and the experiments were subsequently conducted on the trained model through their publicly available docker.

The BSA model \citep{Nguyen:2024} predicts a 3D aging map at the voxel level and subsequently computes specific brain structure ages using a 3D segmentation mask. The brain age of the structures is then used as a biomarker for various downstream applications, such as disease classification. The BSA model was trained on 2,887 T1w brain MRIs from various sources, which is nearly 3 times of our training data. Here, ADNI is the only overlapping dataset with our study. The BSA model achieved an MAE of 1.88 years on a young population (14.8±9.3 yo) and 3.83 years on an older population (64.2±7.9 yo), achieving great accuracy. The trained model is hosted on the volBrain website (https://volbrain.net), our raw test dataset was processed as per the instructions of the original publication.

\subsection{Evaluation metrics}
To evaluate the accuracy for MRI-based brain age estimation, we used the mean absolute error and coefficient of determination ($R^2$). When comparing the absolute errors between DL models based on the test set, two-sided paired-sample t-tests were conducted. In addition, to help characterize the age regression models for healthy and patient cohorts, we also computed the brain age gap (BAG) as ``prediction - ground truth chronological age". For healthy subjects, BAG measures if the trained model over- or under-estimate the chronological age, while for patients with neurodegenerative disorders, it helps gauge disese-related advanced brain aging. For healthy subjects, we used two-sided one-sample t-tests to verify if the BAG value is different from zero for an age regression model. A difference with p-value below 0.05 is considered statistically significant.

\subsection{Grad-RAM for Brain Aging Analysis}

While Grad-CAM \citep{gradcam} is traditionally used for generating saliency maps to help explain DL model's decision in image classification, Grad-RAM extends this approach to handle continuous outputs in regression tasks, including brain age estimation. In this case, Grad-RAM produces a heatmap that highlights the areas of the input MRI that most influences the regression output. In this implementation, a ReLU operation on the gradient activations allows the heatmap to solely represent areas that contribute to increasing the age values. Compared to Grad-CAM, the Grad-RAM heatmaps could vary more per individuals; averaging the heatmaps of specific age groups can better delineate the most important anatomical regions relevant to their aging characteristics. 

To reveal the insights provided by the proposed DL models, Grad-RAM is performed on all test dataset subjects to obtain relevant saliency maps (normalized to the value range of [0,1]). Averaged saliency maps were computed for different age groups in the MNI-ICMB152 space to visualize the relevant anatomical regions that are closely related to aging at different stages of the lifespan. To facilitate anatomy-wise analysis, the Automated Anatomical Labeling (AAL)-16 atlas \citep{AAL} was used to compute the mean Grad-RAM heatmap scores within each of the brain parcellations. Regions with mean activation values over 0.80 are considered to be relevant to aging. Averaged Grad-RAM heatmaps and the associated parcellation-based quantitative values were obtained for the following four subgroups of our testing data: 20-40 yo, 40-60 yo, 60-80 yo, and above 80 yo. To further inspect the potential sex-based differences in brain aging \citep{Armstrong:2019}, we performed a similar analysis by re-grouping and averaging the Grad-RAM heatmaps with respect to sex. Note that the Grad-RAM results of both the RNC-trained and best performing end-to-end training ResNet models were produced for comparison, to demonstrate the potential advantages of contrastive learning approaches.

\subsection{Analysis of Brain Age for Neurodegenerative Conditions}
In neurodegenerative disorders, the patient's brain age gap between biological and chronological aging and disease severity could be correlated \citep{Eickhoff:2021,lowe2016,beheshti2018}, and may serve as an surrogate indicator for disease progression. To explore the application of our proposed method in this domain, we curated groups of AD and PD patients with 3T T1w MRIs from the respective ADNI and PPMI datasets for two different types of analyses. Note that for both cohorts, we compared the proposed RNC-trained and the best end-to-end trained ResNet models that were based on the 1x1x1 $mm^{3}$ high-resolution MRIs. 

\textbf{In the first analysis}, we estimated the biological brain age using the ResNet50-highRez model from conventional training and RNC-highRez model (RezNet18 backbone) for the patient groups, and then we computed the correlation between the brain age gap with disease severity metrics. Specifically, for Alzheimer's disease, we used the ADAS-Cog-11 score \citep{adas}, which evaluates memory, language, praxis, and orientation, and is widely adopted for assessing cognitive function in AD patients. On the other hand, for Parkinson's disease, we used the UPDRSIII score \citep{updrs} that focuses specifically on motor symptoms, the primary clinical manifestations of PD, and is a more objective metric compared to other self-evaluation-based scores of the UPDRS assessment. \textbf{In the second analysis}, we obtained population-averaged Grad-RAM heatmaps based on DL-based age estimation using the proposed RNC method to reveal the relevant anatomical patterns that are associated with aging under the influence of neurodegenerative disorders in comparison to healthy aging.


\section{Results}
\subsection{Evaluation of Deep Learning Models}
\subsubsection{Brain Age Estimation Accuracy}
\begin{figure}
    \centering
    \includegraphics[width=0.8\linewidth]{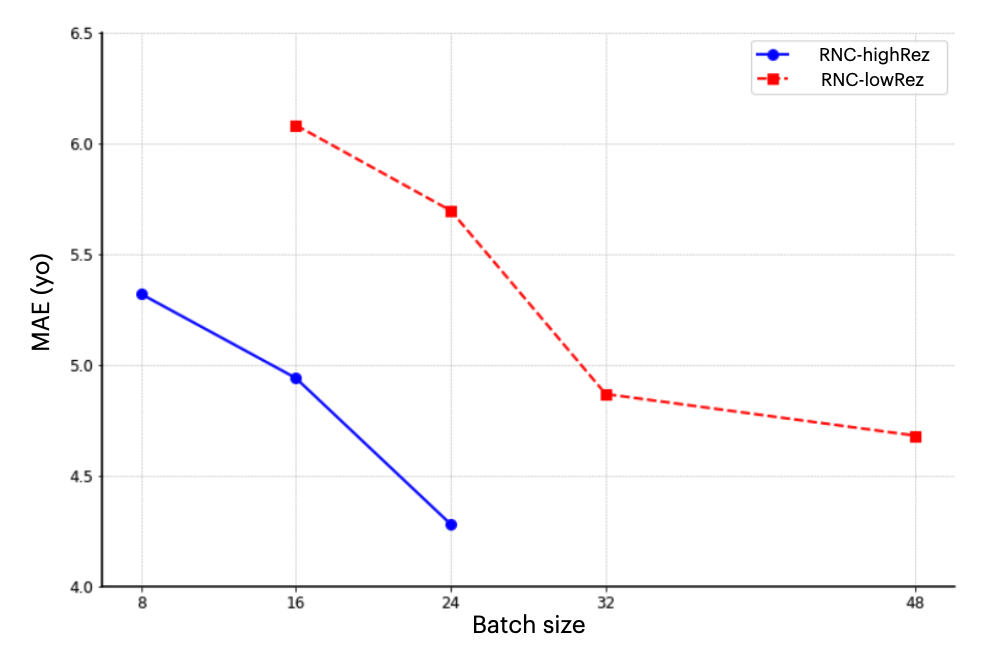}
    \caption{Investigation of different batch sizes on MAE for RNC models trained on 1x1x1 $mm^{3}$  resolution and 2x2x2 $mm^{3}$ resolution data.}
    \label{fig:RegressionBatchSize}
\end{figure}

The results of the proposed contrastive learning method and the variants of the ResNets models based on end-to-end training for MRI-based age estimation are shown in Table \ref{table:results1}, where results from both the high-resolution (1x1x1 $mm^{3}$) and low-resolution (2x2x2 $mm^{3}$) settings are compared. In addition, we also compared the performance of the ResNet50 model with and without data augmentation during training. First, when examining the difference between the ResNet50 variants with and without data augmentation, we observed an evident performance boost from data augmentation, particularly at the 1x1x1 $mm^{3}$ resolution. Secondly, with data augmentation, the depth of 50 for the ResNets has shown better performance compared to both the deeper 101-layer and shallower 18-layer versions. Notably, increasing the data resolution from 2x2x2 $mm^{3}$ to 1x1x1 $mm^{3}$ significantly enhanced model performance across all ResNet models ($p<0.05$). Lastly, the RNC-trained models exhibited the best performance across both resolutions, with the 1x1x1 $mm^{3}$ resolution version (RNC-highRez) outperforming all other models across all metrics with an MAE of 4.27±3.32 years and $R^2$ of 0.93. As a compromise to the GPU memory limitation, we used ResNet18 for RNC-trained high-resolution model (RNC-highRez) than ResNet50 for RNC-lowRez. Thus, the MAE and $R^2$ improvements are limited at the higher resolution setting. In addition, as shown in Fig. \ref{fig:RegressionBatchSize}, the batch size had an important impact on the RNC model. The results with the largest possible batch size without a GPU memory issue are presented in Table \ref{table:results1}, with 48 for 2x2x2 $mm^{3}$ and 24 for 1x1x1 $mm^{3}$. Based on this investigation, we used ResNet50-highRez and RNC-highRez for the rest of our experiments and analyses.

\begin{table}[bt]
\caption{Accuracy assessments of MRI-based brain age estimation for end-to-end trained ResNet variants and RNC models under different image resolutions and data augmentation conditions. * Indicates $p<0.01$.}
\label{table:results1}
\centering
\renewcommand{\arraystretch}{1.2} 
\setlength{\tabcolsep}{10pt} 
\begin{threeparttable}
  \resizebox{\textwidth}{!}{
\begin{tabular}{l|cc}
\toprule
\textbf{Model} & \textbf{MAE±Std(yr)} & \textbf{\textit{R$^2$}} \\
\midrule
\multicolumn{3}{c}{\small{1x1x1 $mm^3$ Resolution}} \\ 
\midrule
ResNet50 w/o Data Augmentation & $6.19 \pm 5.23$ & 0.73* \\
ResNet18-highRez & $5.21 \pm 4.22$ & 0.86* \\
ResNet50-highRez & $4.92 \pm 3.96$ & 0.90* \\
ResNet101-highRez & $5.85 \pm 5.05$ & 0.85* \\
RNC-highRez (ResNet18 backbone) & $4.27 \pm 3.32$& 0.93* \\
\midrule
\multicolumn{3}{c}{\small{2x2x2 $mm^3$ Resolution}} \\ 
\midrule
ResNet50 w/o Data Augmentation & $6.42 \pm 6.12$ & 0.71* \\
ResNet18-lowRez & $6.12 \pm 5.60$ & 0.72* \\
ResNet50-lowRez & $5.47 \pm 4.90$ & 0.80* \\
ResNet101-lowRez & $6.76 \pm 5.89$&0.73* \\
RNC-lowRez (ResNet50 backbone) & $4.68 \pm 3.80$& 0.90* \\
\bottomrule
\end{tabular}}
\end{threeparttable}
\end{table}

\subsubsection{Comparison with the State-of-the-Art Methods}
Table \ref{table:results2} compares the accuracy of the best performing ResNet50-highRez and RNC-highRez against the two benchmark models, SFCN-reg and BSA. The SFCN-reg model had a significantly higher MAE of 5.46±3.89 years ($p=0.002$) and a lower $R^2$ of 0.85 than our proposed RNC-highRez model, despite being trained on data that is $\sim$42 times of our training set. On the other hand, the BSA model \citep{Nguyen:2024} obtained a significantly better MAE at 4.08±3.23 years ($p<0.01$), but a lower $R^2$ than RNC-highRez. However, when inspecting the estimation accuracy for the older population group over 60 yo (N=62), RNC-highRez offered an MAE of 4.30±3.23 years and $R^2$ of 0.77 while BSA resulted in slightly inferior MAE of 4.36±3.44 years and $R^2$ of 0.74. In addition, the mean BAG of SFCN-reg is lower than the other methods.

\begin{table}[ht]
    \centering
    \caption{Results of our best-performing end-to-end trained ResNet50-highRez model and proposed RNC-highRez model, in comparison to the pretrained BSA and SFCN-reg models. * Indicates $p<0.05$.}
    \label{table:results2}
    \setlength{\tabcolsep}{6pt}
    \resizebox{\textwidth}{!}{
    \begin{tabular}{l|c|c|c|c}
        \toprule
        \textbf{Model} & \textbf{Train Data} & \textbf{MAE±Std(yr)} & \textbf{Mean BAG (yr)} & \textbf{R\textsuperscript{2}} \\
        \midrule
        ResNet50-highRez & 1294    & $4.92 \pm 3.96$ & -0.35 & 0.90* \\
        RNC-highRez      & 1294    & $4.27 \pm 3.32$ & -0.61 & 0.93* \\
        BSA              & $\sim$2887 & $4.08 \pm 3.23$ & -0.61 & 0.89* \\
        SFCN-reg         & 42,829  & $5.46 \pm 3.89$ & -3.21 & 0.85* \\
        \bottomrule
    \end{tabular}
    }
\end{table}

\subsection{Grad-RAM Analysis of Healthy Population}
\begin{figure}
    \centering
    \includegraphics[width=1\linewidth]{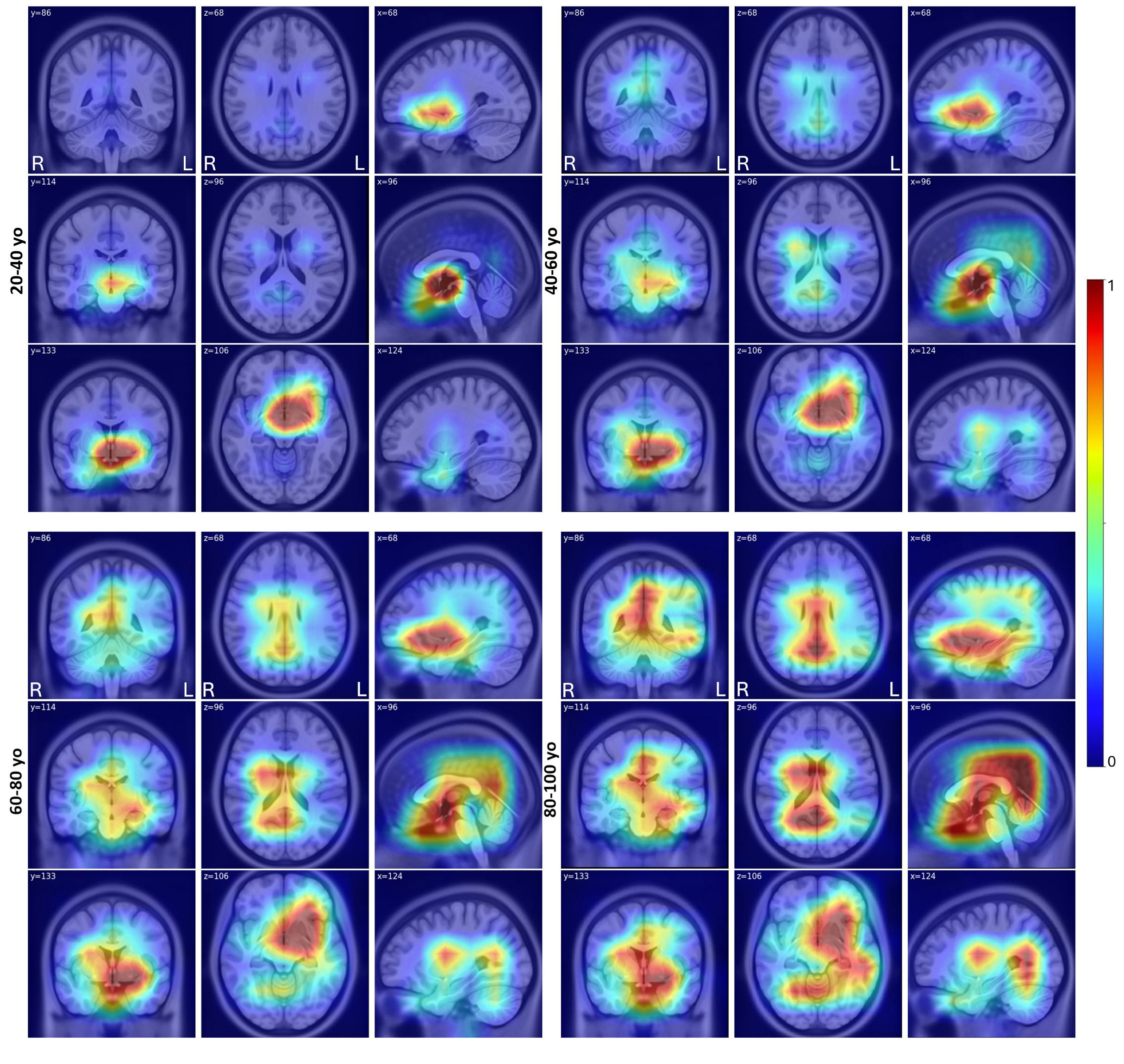}
    \caption{Averaged Grad-RAM heatmaps produced from the ResNet 50-highRez model based on the healthy control test set for the age groups of 20-40 yo, 40-60 yo, 60-80 yo, and 80-100 yo.}
    \label{fig:Grad-RAMresnet_updated2}
\end{figure}

As an exploratory study, we computed the mean values from the population-averaged Grad-RAM heatmaps within 116 brain parcellations based on the AAL116 atlas. Upon inspecting the population-averaged Grad-RAM results across different age groups, the ResNet50-highRez model highlighted fairly consistent regions of interest across the age groups, with virtually no differences between sexes. Specifically, for all age groups, the main areas the model focuses on are in the left hemisphere (in order of decreasing magnitude, with the mean saliency greater than 0.8): pallidum, amygdala, and putamen. The younger population (20$\sim$40 yo) has a high saliency concentration in the left hippocampus and pallidum. As the population ages, several other areas become more involved, especially for the population above 80 yo, and these include the right pallidum, thalamus, caudate, left parahippocampal gyrus, left hippocampus, posterior cingulum, vermis, and left frontal orbital gyrus. Based on the observation, with aging, the ResNet50-highRez model focuses on wider and more diverse regions of interest while the primary focus is still on subcortical regions. A visual inspection of the averaged Grad-RAMs (see Fig. \ref{fig:Grad-RAMresnet_updated2}) further confirms our results, showing the younger population (20$\sim$40 yo) having a more concentrated activation area, and the older population expanding the activation area in the subcortical regions, ventricles, and the cortex.

\begin{figure}
    \centering
    \includegraphics[width=1\linewidth]{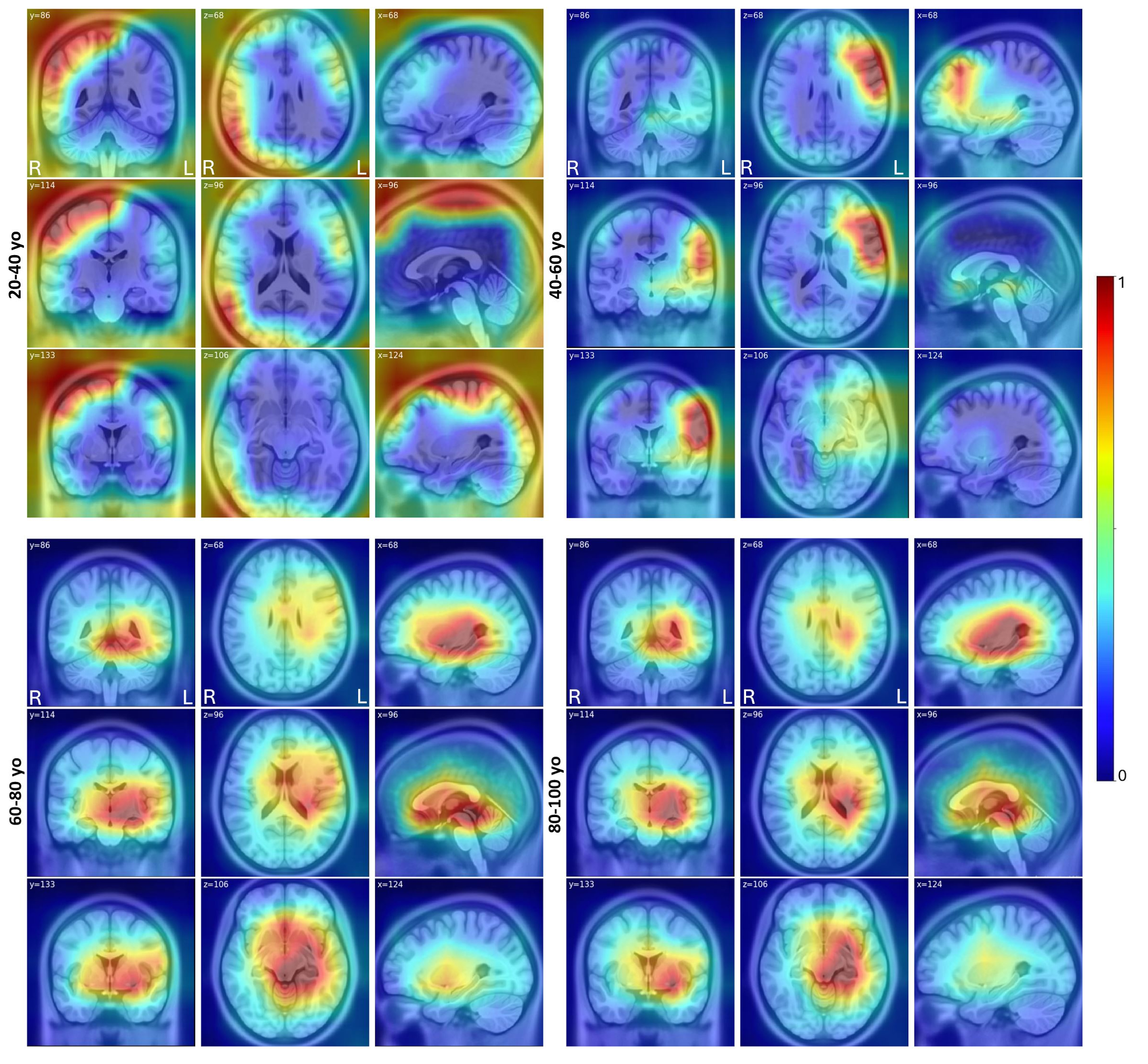}
    \caption{Averaged Grad-RAM heatmaps produced from the RNC-highRez model based on the healthy control test set for the age groups of 20-40 yo, 40-60 yo, 60-80 yo, and 80-100 yo.}
    \label{fig:Grad-RAMcontrastive}
\end{figure}

In contrast to ResNet50-highRez, the RNC-highRez model showed more distinct Grad-RAM activation patterns across different age groups. For the 20$\sim$40 yo group, the activated areas from the DL model are on the right parietal and lateral frontal lobe, including (in the order of decreasing magnitude, with the mean saliency greater than 0.8) inferior parietal, right postcentral, precentral, supramarginal, and angular gyrus, and right paracentral lobule. For the 40$\sim$60 yo group, the model shifts the attention with a focus towards the left lateral frontal lobe, particularly the frontal inferior operculum, frontal inferior triangularis, rolandic operculum, and with a broader and relatively weaker focus on areas around it, such as the left insula. Finally, the 60$\sim$80 yo group and the 80+ yo group have highly similar activation patterns, focusing on subcortical regions and to a lesser extent, the left lateral frontal lobe. The main regions of interest remain on the left hemisphere, including (in the order of decreasing magnitude, with the mean saliency greater than 0.8) the thalamus, pallidum, putamen, hippocampus, amygdala, caudate, and vermis. A visual inspection (see Fig. \ref{fig:Grad-RAMcontrastive}) confirms these findings from the parcellation-wise mean Grad-RAM values and suggests that the younger population (20$\sim$40 yo) has more activation in the lateral frontal and parietal lobes, shifting towards the left hemisphere's subcortical regions and lateral ventricles along with the aging process.

Similar to the case of ResNet50-highRez, the RNC-highRez model also did not demonstrate drastic Grad-RAM pattern differences between male and female aging although some nuanced distinctions in the strengths of activation were present for across age groups and within the same age groups. On average, the right parietal inferior lobe, right angular gyrus, right paracentral lobule, right parietal superior lobe, and left frontal superior lobe, are about 40\% more activated for males, with the contribution mainly from the 20$\sim$40 group. 

\subsection{Brain Age Estimation for Alzheimer's and Parkinson's Disease Patients}
\begin{figure}
    \centering
    \includegraphics[width=1\linewidth]{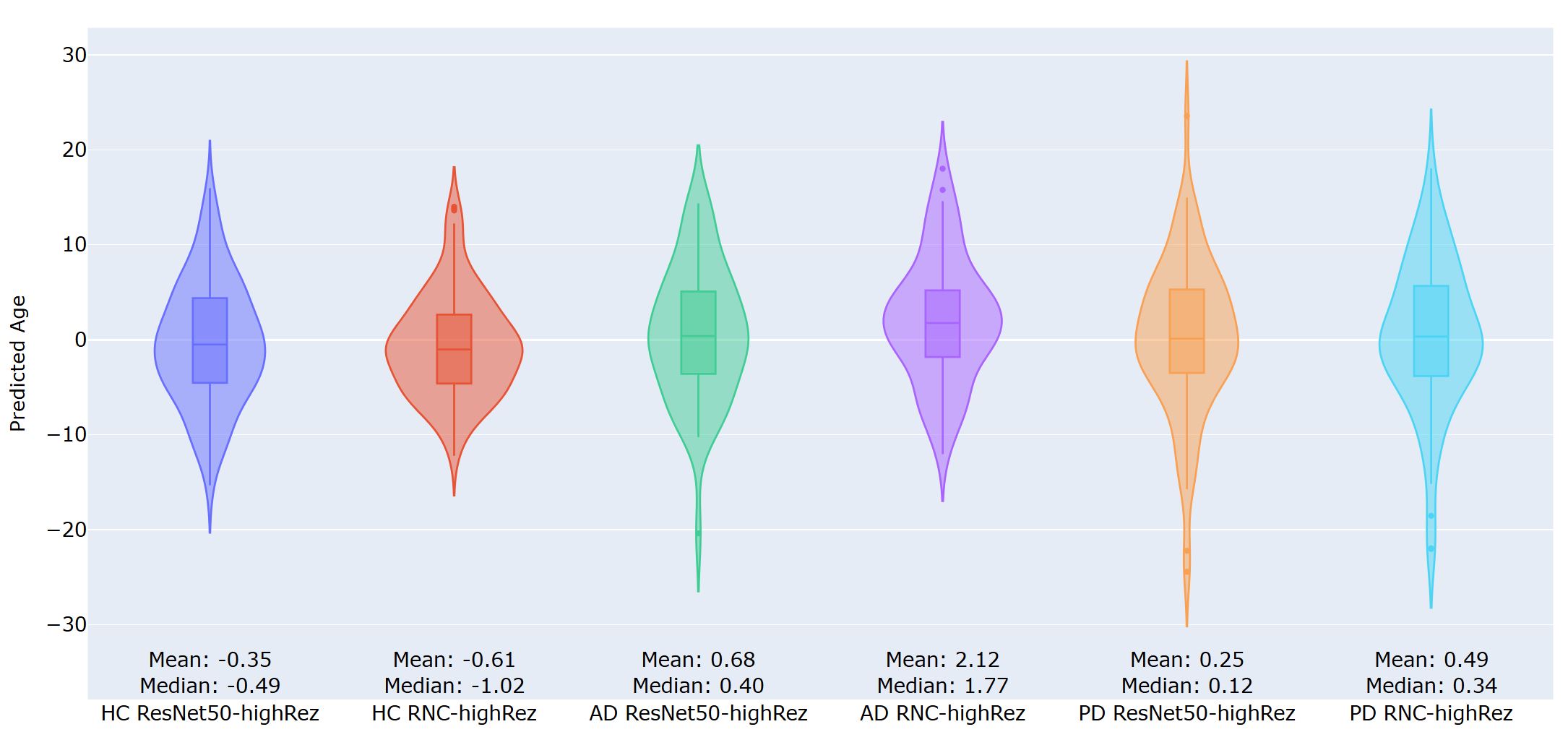}
    \caption{Violin plot of differences between the estimated brain age and the ground truths (BAGs) for the ResNet50-highRez and RNC-highRez models on healthy and diseased populations. }
    \label{fig:violinplot}
\end{figure}

In Table \ref{table:resultsADPD}, we list the age estimation accuracy of ResNet50-highRez and RNC-highRez in terms of the MAE and mean brain age gap values, as well as ``brain age gap vs. disease severity" correlations for the curated AD and PD patient cohorts. In both disease groups and between the ResNet50-highRez and RNC-highRez models, the MAE and the associated standard deviation exhibit an increase compared to healthy subjects, as anticipated in diseased patients. In addition, the mean brain age gap for the diseased population demonstrated an increase compared to healthy controls on average, as illustrated in Fig. \ref{fig:violinplot}. As expected, the healthy control group exhibits a mean BAG closest to 0, with the ResNet50-highRez and RNC-highRez model showing a discrepancy of -0.35 years and -0.61 years, respectively. In contrast, the AD group shows a substantial increase in mean BAG to +0.68 years for the ResNet50-highRez ($p < 0.01$) and +2.12 years for the RNC-highRez model, which is significantly higher than that of the ResNet50-highRez ($p < 0.01$). We also observed that the brain age gaps are correlated to ADAS-Cog-11 scores in the AD cohort, with the RNC-highRez model achieving a higher correlation of 0.37 ($p=0.009$). In addition, the RNC-highRez model predicts older ages than the ResNet50-highRez counterpart on average. In the PD cohort, the ResNet50-highRez model exhibits a modest increase in mean BAG compared to the healthy controls without a statistically significant difference, whereas the RNC-highRez model shows a significantly higher mean BAG of +0.49 ($p < 0.01$).  No significant correlation was found between the brain age gap and UPDRSIII scores, with correlations of 0.030 ($p$=0.80) for the ResNet50-highRez and 0.045 ($p$=0.63)  for the RNC-highRez.

\begin{table}[ht] 
\centering
\caption{Age estimation discrepancy in Alzheimer's and Parkinson's disease cohorts measured with MAE and brain age gap (BAG) in years, as well as the Pearson correlation between brain age gap and disease severity metrics for AD (ADAS-Cog-11) and PD (UPDRSIII). * Indicates mean BAG significantly different from zero ($p<0.05$). $\star$ indicates significant correlations ($p<0.05$).}
\label{table:resultsADPD}
\begin{threeparttable}
\renewcommand{\arraystretch}{1.2} 
\resizebox{\textwidth}{!}{ 
\begin{tabular}{|c|c|c|c|c|} 
\hline 
\textbf{Population} & \textbf{Model} & \textbf{MAE ± Std. Dev.} & \textbf{BAG (mean ± Std. Dev.)} & \textbf{Corr. BAG vs. Disease Severity}\\ 
\hline 
\multirow{2}{*}{AD ($n=80$)} & ResNet50-highRez &  5.49 ± 4.42 &  + 0.68 ± 7.00 &  0.25$\star$\\ 
& RNC-highRez & 5.61 ± 4.40 & + 2.12 ± 6.71* & 0.37$\star$\\ 
\hline 
\multirow{2}{*}{PD ($n=61$)} & ResNet50-highRez &  6.01 ± 5.35 &  + 0.25 ± 8.10 &  0.030\\ 
& RNC-highRez & 5.85 ± 5.21 & + 0.49 ± 7.90 & 0.045\\ 
\hline
\end{tabular}
}
\end{threeparttable}
\end{table}

\subsection{Grad-RAM Analysis for Alzheimer's and Parkinson's Disease Patients}

\begin{figure}
    \centering
    \includegraphics[width=1\linewidth]{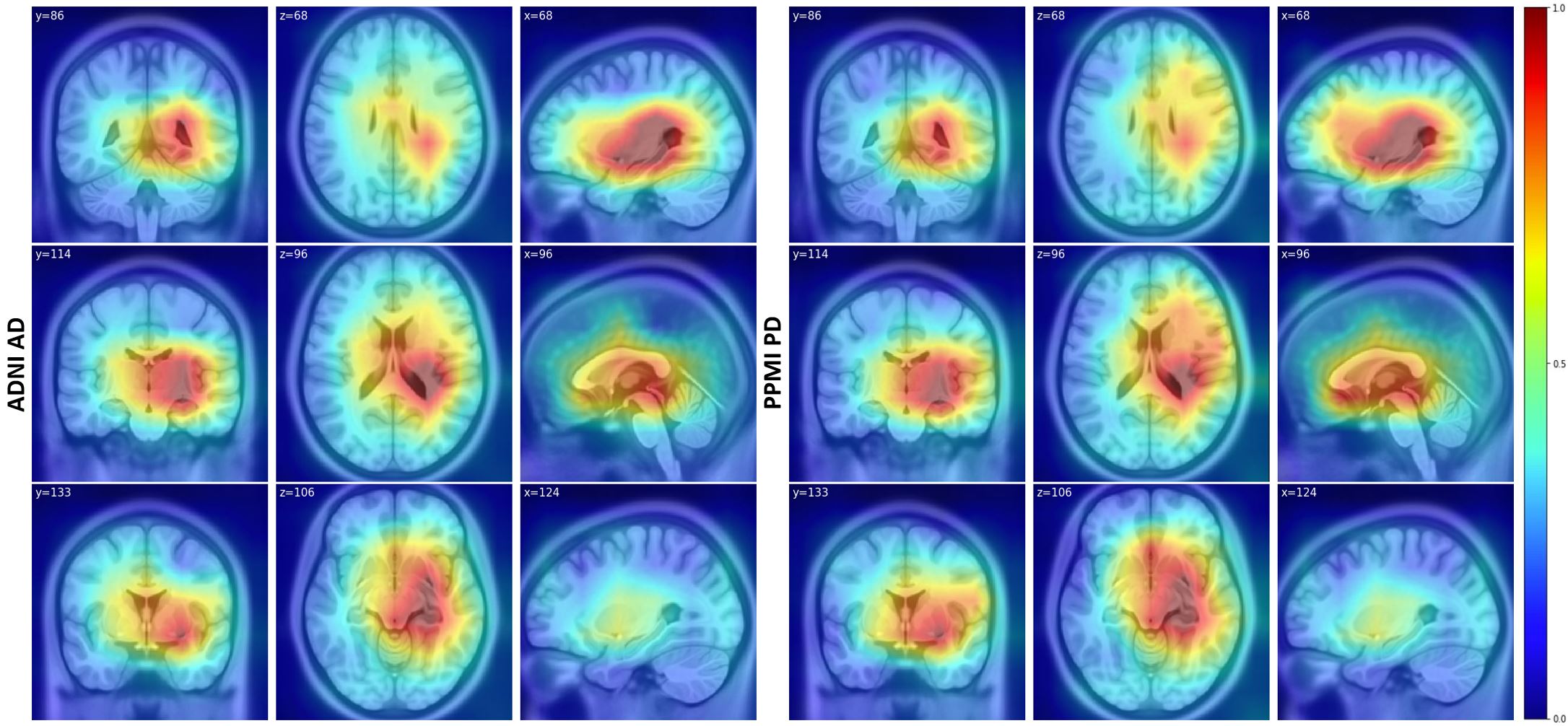}
    \caption{Averaged Grad-RAM heatmaps produced from the RNC-highRez model for the AD (left) and PD (right) patients.}
    \label{fig:GradRAM-ADPD}
\end{figure}

To inspect the potential discrepancy of the aging patterns in the AD and PD population's anatomy from that of the healthy aging one, we generated the Grad-RAM heatmaps with the RNC-highRez model for the AD and PD groups, and we employed the AAL116 atlas to investigate the relevant anatomical regions similar to Section 3.2. The Grad-RAM outputs are shown in Fig. \ref{fig:GradRAM-ADPD}.

For \textbf{AD patients}, our proposed RNC-highRez model revealed similar activation patterns to the healthy controls of the same age range (60$\sim$80 yo and 80+ yo), which are concentrated on the left hemisphere (in the order of decreasing mean Grad-RAM values, with the mean saliency greater than 0.8): thalamus, pallidum, putamen, hippocampus, amygdala, and caudate. Compared to the Grad-RAM heatmap of the healthy controls with the same model, there are slight increases in activation in the left temporal and lateral frontal lobe, caudate, and putamen. Visually, the hippocampus, left lateral ventricle and midbrain are more activated.

For \textbf{PD patients}, with the RNC-highRez model, the main activation patterns remain similar to those of the HC group for the same age range (60$\sim$80 yo and 80+ yo), once again on the left hemisphere (in the order of decreasing mean Grad-RAM values, with the mean saliency greater than 0.8):  thalamus, pallidum, putamen, hippocampus, amygdala, and caudate. There are some increases in the activation level to the left temporal and lateral frontal lobe compared to the HC cohort. Visually, the left lateral ventricle, basal ganglia, and midbrain are more activated.

\section{Discussion}

\subsection{Deep Learning Model Performance}

The absence of a common benchmarking dataset for brain age estimation introduces great variability in accuracy assessment across different studies, as the reported accuracy can depend on dataset size, imaging properties (e.g., scanner types and imaging protocols), and subject age distribution. Notably, studies with DL models trained on older cohorts (over 25 yo) reported higher MAEs compared to those that also include children and adolescent populations (0-18yo) \citep{Nguyen:2024, DLbrainagepredRev}, who have more pronounced anatomical development. Consequently, it is important to well consider these factors when comparing the accuracies of different MRI-based age estimation models reported in the literature. When applying the SOTA models trained on significantly larger datasets to our testing dataset, we observed an under-performance compared to their expected out-of-domain accuracy using the same data pre-processing procedures. Specifically, BSA \citep{Nguyen:2024} trained on roughly 2,887 subjects achieved an MAE of 4.08 years with our test set compared to their reported out-of-domain results of 3.83 years, although they provided an improvement in $R^2$ of 0.89 compared to their reported 0.62. Similarly, SFCN-reg \citep{pyment} trained on 42,829 T1w MRIs achieved an MAE of 5.46 years compared to their reported MAE of 3.90 years, and $R^2$ of 0.89 instead of the reported 0.94. In terms of the technical approach, our proposed method is more related to SFCN-reg while the more complex BSA was constructed based on 125 3D UNets for voxel-wise brain age estimation, an image segmentor, and a support vector regressor for global age estimation. As reported in the work of \cite{SFCN}, a SFCN variant that adopted a ResNet-50 trained on 12,949 subjects (44-80 yo) with similar data augmentation methods to ours, achieved an MAE of 2.32 years on their in-domain test set. Therefore, the comparatively smaller training dataset and demographic differences may contribute to the lower MAEs observed with our proposed method compared to the best state-of-the-art deep learning model. 

Our proposed RNC-highRez model significantly outperformed all our end-to-end-trained ResNet variants and the baseline SFCN-reg model, and had comparable performance to the BSA model. While further enrichment of the dataset can be beneficial \citep{DLbrainagepredRev, Nguyen:2024, SFCN} to enhance the accuracy, our experiments have well demonstrated the robustness and data efficiency of the RNC approach in 3D medical image applications, which are consistent with the conclusion of the original work in natural vision tasks \citep{zha2023rankncontrast}. A few factors were found to greatly impact performance of the proposed RNC-highRez model. Besides data augmentation, increasing the batch size for the RNC model improved performance, as noted by previous literature on supervised contrastive learning \citep{supconlossgoogle}. Additionally, we observed a positive impact of finer brain MRI resolution on the age estimation, suggesting that structural changes related to aging can be more nuanced. However, due to the hardware constraints and the large volumetric inputs, a balance has to be made between the feature encoder model size and the training batch size. This poses the main challenge in contrastive learning for volumetric data, and we are the first to explore rank-based contrastive learning in regression tasks for 3D medical image applications.

\subsection{Grad-RAM saliency maps in healthy subjects} 
Our Grad-RAM analyses provided insights into the brain regions that are most associated with healthy aging by the selected DL models across different age groups. The saliency map results echo with the existing literature while notable differences exist between the two different model training approaches: ResNet50-highRez and RNC-highRez.

The ResNet50-highRez model present fairly consistent areas of interest in the Grad-RAM result across different age groups, focusing primarily on the limbic system, basal ganglia, cortices, and ventricles. For MRI-based age estimation, the model mainly focuses on subcortical regions, which are known to have significant volume and morphological changes through aging \citep{Fjell2010, Fujita:2023, convit1995, Clifford1992}. The saliency of the orbital cortex also resonates with known aging-related cortical thinning and ventricular enlargement \citep{Peters:2006, Fjell2009, Fujita:2023}. On the other hand, the RNC-highRez model's saliency maps showed more distinct patterns with regions of interest varying significantly with age. For younger populations (20$\sim$40 yo), the model focused on the right parietal and lateral frontal lobes, shifting towards the left lateral frontal lobe in the 40$\sim$60 yo age group. For older groups (60+ yo), the focus transitioned to subcortical regions similar to those identified by the ResNet50-highRez counterpart, showing once again influence on the limbic system and basal ganglia from aging, but with the notable addition of the cerebellum. These observations corroborate with existing literature, which reported volume reduction to the limbic system and basal ganglia \citep{Mu1999, Clifford1992} and ventricle enlargement \citep{Fujita:2023, Peters:2006}. 

Other studies using various CNN-based explainability methods \citep{Yin:2023, Joo:2023} also find similar activation patterns in the subcortical regions, and also reported activations on the cortices compared to our study. Aging results in substantial thinning of cortical structures \citep{Gunning-Dixon:2009, Peters:2006} which is often the focus of most existing studies using the popular voxel-based morphometry (VBM) \citep{Han2022,liem2016}. In contrast, our DL-based analysis revealed heavier weights on the sub-cortical regions in aging, which is consistent with the study of \cite{HEPP2021101967}.  The RNC-highRez model also showed slight sex-based differences, particularly in younger populations, similar to  \cite{Yin:2023}. However, the impact of sex is not strong in our analysis and may be a result of a smaller sample size and the differences in explainability methods.

\subsection{Alzheimer's and Parkinson's Disease}
\label{SubSec.:ADandPD}

\subsubsection{Brain Age Estimation for Alzheimer's and Parkinson's Disease Patients}

Our findings reveal elevated MAEs and mean brain age gaps in both Alzheimer's and Parkinson's disease groups compared to healthy controls. Specifically, the AD cohort has a mean brain age gap of +0.68 years from the ResNet50-highRez model and a larger one of +2.12 years from the RNC-highRez model. These results, especially with the RNC-highRez model are similar to those previously reported by \cite{Nguyen:2024} of +3.4 years and \cite{Sendi:2021} of +3.3 years . The significant difference of BAG ($p < 0.01$) between the ResNet50-highRez and RNC-highRez models suggest that the RNC-highRez model may be more sensitive to nuanced anatomical changes due to AD. On the other hand, for the PD cohort, the ResNet50-highRez model shows no significant discrepancy between estimate and chronological age, whereas the RNC-highRez model detects a significantly elevated mean brain age gap of +0.49 years ($p < 0.01$). This is similar to the result of +1.0 years reported by \cite{Nguyen:2024}, which is weaker than the larger positive gap reported by Eickhoff et al. of +2.8 years \citep{Eickhoff:2021}. For AD and PD cohorts, the higher senstivity in detection the disease-related advanced brain aging confirms the robustness of the employed contrastive learning strategy.

When associating the brain age gaps with the ADAS-Cog-11 scores in AD, we found a moderate correlation of 0.37 ($p=0.009$) with the RNC-highRez model. Our result is comparable to that of \cite{lowe2016} ranging between 0.25 to 0.56, and slightly higher than the correlation of 0.26 obtained by \cite{beheshti2018}. Conversely, there is no statistically significant correlation between the brain age gap and UPDRSIII scores in our tested PD cohort. This is similar to the conclusion of \cite{chen2024}, but \cite{Eickhoff:2021} found a significant yet weak correlation of 0.14 to 0.17 . This potentially suggests that anatomical alterations due to PD are relatively subtle in the presence of natural brain aging \citep{Nguyen:2024, Mishra:2019}. To detect these more nuanced PD-induced brain structural degeneration, ultra-high-field MRIs (e.g., 7T) that allow more detailed and sharper anatomical depiction, particularly for the subcortical structures could potentially be instrumental \citep{7TPD}. However, these types of MRI scanners also pose challenges, such as high cost, less accessibility, and can tend to have large distortions in some brain regions. Finally, the discrepancies between our correlation measures and those previously reported are also subject to the sampling of the patients, as well as the size of the cohorts under study.

\subsubsection{Grad-RAM Analysis for Alzheimer's and Parkinson's Disease Patients}
The Grad-RAM heatmap from the RNC-highRez model on the AD cohort showed similar regions of interest related to aging to those on healthy controls in the matching age groups, focusing on the left hemisphere's limbic system, lateral ventricle, and basal ganglia. However, higher saliency values in AD than in HC are observed in the left temporal and lateral frontal lobe, hippocampus, caudate, and putamen. The hippocampus and surrounding areas present themselves as the most relevant region for aging. This mirrors the common concept of significant hippocampal atrophy during AD progression \citep{atrophyAD, frisoni2010, Mu2011}.  Additionally, \cite{Yin:2023}, who also used saliency maps from a CNN found similar dependence on the limbic system for MCI and AD cohorts. In the voxel-based brain age estimation by \cite{Nguyen:2024}, they found more advanced brain aging in the hippocampus and amygdala than the rest of the brain while \cite{Gianchandani:2024voxellevelapproach} reported the caudate, insula and putamen to be the most prominent in brain aging. These findings further reinforce the potential importance of the limbic system, especially hippocampus and the basal ganglia on disease progression of AD that can induce accelerated biological brain aging.

In terms of the PD cohort, the population-averaged saliency map from Grad-RAM also demonstrated similar patterns with respect to age estimation to the healthy population in the same age groups, highlighting the left hemisphere's thalamus, pallidum, putamen, hippocampus, amygdala, and caudate. Although the basal ganglia and limbic system are known to undergo morphological changes in PD \citep{prakash2016,chen2024,pieperhoff2022}, the prominence of frontal and temporal cortices noted in previous studies \citep{chen2024, pieperhoff2022} was not observed in our saliency maps. Conversely, some studies also reported no areas that have significant morphological changes associated with PD in terms of accelerated aging patterns \citep{Nguyen:2024} and alternative metrics that characterize tissue microstructures may be more sensitive \citep{Xiao2021}.

In general, the Grad-RAM results from the RNC-highRez model demonstrated higher saliency values in regions affected by AD and PD compared to those of the ResNet50-highRez counterpart. This observation suggests that RNC-highRez may be more adapted at identifying disease-specific features within our sample cohort. However, strong similarities are still present in Grad-RAM saliency patterns between AD/PD and HC, and the distinct aging-related and disease-induced changes may be entangled in the saliency patterns that were produced. This warrants further investigation to determine whether Grad-RAM can reliably identify regions of accelerated aging for neurodegenerative disease samples with high specificity. 

\section{Conclusion}
We have proposed the RNC-HighRez model, which adapts the Rank-N-Contrast loss function in a contrastive learning strategy for brain age estimation based on 3D MRI for the first time. The proposed method has significantly outperformed multiple end-to-end trained ResNet variants in brain age estimation across multiple metrics, and has shown better or similar performance against existing DL models trained with much larger datasets. By using Grad-RAM on the proposed technique, which visually explains the decision-making process of the algorithm, we highlight anatomical regions that are known to be associated with brain aging, and the proposed RNC-highRez model did so with more nuance than the ResNet counterpart. For neurodegenerative diseases, we demonstrated that our DL techniques can detect accelerated aging in diseased brains that is associated with disease severity, with the model trained using contrastive learning offering more robust detection. 

\section*{Data Acknowledgment}
\noindent
Data used in the preparation of this article was obtained on 2024-03-21 from the Parkinson’s Progressive Markers Initiative (PPMI) database, RRID: SCR\_006431. For up-to-date information on the study, visit www.ppmi-info.org. 

PPMI – a public-private partnership – is funded by the Michael J. Fox Foundation for Parkinson's Research and funding partners, including 4D Pharma, Abbvie, AcureX, Allergan, Amathus Therapeutics, Aligning Science Across Parkinson’s, AskBio, Avid Radiopharmaceuticals, BIAL, BioArctic, Biogen, Biohaven, BioLegend, BlueRock Therapeutics, Bristol-Myers Squibb, Calico Labs, Capsida Biotherapeutics, Celgene, Cerevel Therapeutics, Coave Therapeutics, DaCapo Brainscience, Denali, Edmond J. Safra Foundation, Eli Lilly, Gain Therapeutics, GE HealthCare, Genentech, GSK, Golub Capital, Handl Therapeutics, Insitro, Jazz Pharmaceuticals, Johnson \& Johnson Innovative Medicine, Lundbeck, Merck, Meso Scale Discovery, Mission Therapeutics, Neurocrine Biosciences, Neuron23, Neuropore, Pfizer, Piramal, Prevail Therapeutics, Roche, Sanofi, Servier, Sun Pharma Advanced Research Company, Takeda, Teva, UCB, Vanqua Bio, Verily, Voyager Therapeutics, the Weston Family Foundation and Yumanity Therapeutics.

The Alzheimer's disease MRI data in this study were obtained from the Alzheimer’s Disease Neuroimaging Initiative (ADNI) database, which is available to all researchers.




\section*{Funding}
\noindent
Y.X. is supported by the Fond de la Recherche du Québec – Santé (FRQS-chercheur boursier Junior 1) and Parkinson Quebec. M.K-O. is supported by the Fond de la Recherche du Québec – Santé (FRQS-chercheur boursier Junior 2).

\section*{Declaration of Competing Interests}
\noindent
The authors declare no competing interests.


\bibliographystyle{elsarticle-harv} 
\bibliography{cas-refs}

\end{document}